\documentclass[oribibl]{llncs}

\usepackage{amsmath}
\usepackage{graphicx}
\usepackage{psfrag}

\title{Probabilistic Search for Object Segmentation and Recognition}

\author{Ulrich Hillenbrand \and Gerd Hirzinger}

\institute{Institute of Robotics and Mechatronics, German Aerospace Center\\Oberpfaffenhofen, 82234 Wessling, Germany\\\email{Ulrich.Hillenbrand@dlr.de}}

\begin{document}

\maketitle

\begin{abstract}
The problem of searching for a model-based scene interpretation is analyzed within a probabilistic framework. Object models are formulated as generative models for range data of the scene. A new statistical criterion, the truncated object probability, is introduced to infer an optimal sequence of object hypotheses to be evaluated for their match to the data. The truncated probability is partly determined by prior knowledge of the objects and partly learned from data. Some experiments on sequence quality and object segmentation and recognition from stereo data are presented. The article recovers classic concepts from object recognition (grouping, geometric hashing, alignment) from the probabilistic perspective and adds insight into the optimal ordering of object hypotheses for evaluation. Moreover, it introduces point-relation densities, a key component of the truncated probability, as statistical models of local surface shape.
\end{abstract}

\noindent Published in Proceedings {\em European Conference on Computer Vision 2002},\\Lecture Notes in Computer Science Vol.\ 2352, Springer, pp.\ 791--806.

\section{Introduction}
\label{Intro}

Model-based object recognition or, more generally, scene interpretation can be conceptualized as a two-part process: one that generates a sequence of hypotheses on object identities and poses, the other that evaluates them based on the object models. Viewed as an optimization problem, the former is concerned with the search sequence, the latter with the objective function. Usually the evaluation of the objective function is computationally expensive. A reasonable search algorithm will thus arrive at an acceptable hypothesis within a small number of such evaluations.

In this article, we will analyze the search for a scene interpretation from a probabilistic perspective. The object models will be formulated as generative models for range data. For visual analysis of natural scenes, that is, scenes that are cluttered with multiple, non-completely visible objects in an uncontrolled context, it is a highly non-trivial task to optimize the match of a generative model to the data. Local optimization techniques will usually get stuck in meaningless, local optima, while techniques akin to exhaustive search are precluded by time constraints. The critical aspect of many object recognition problems hence concerns the generation of a clever search sequence.

In the probabilistic framework explored here, the problem of optimizing the match of generative object models to data is alleviated by the introduction of another statistical match criterion that is more easily optimized, albeit less reliable in its estimates of object parameters. The new criterion is used to define a sequence of hypotheses for object parameters of decreasing probability, starting with the most probable, while the generative model remains the measure for their evaluation. For an efficient generation of the search sequence, it is desirable to make the new criterion as simple as possible. On the other hand, for obtaining a short search sequence for acceptable object parameters, it is necessary to make the criterion as informative as is feasible. As a key quantity for sequence optimization, an estimate of features' posterior probability enters the new criterion. The method is an alternative to other, often more heuristic strategies aimed at producing a short search sequence, such as checking for model features in the order of the features' prior probability \cite{Goad83,Kuno_etal91}, in a coarse-to-fine hierarchy \cite{Fleuret&Geman99}, or as predicted from the current scene interpretation \cite{Faugeras&Hebert87,Grimson90}.

Classic hypothesize-and-test paradigms have been demonstrated in RANSAC-like \cite{Fischler&Bolls81,Torr&Zisserman00} and alignment techniques \cite{Huttenlocher&Ullman87,Huttenlocher&Ullman90}. The method proposed here is more similar to the latter in that testing of hypotheses is done with respect to the full data rather than on a sparse feature representation. It significantly differs from both, however, in that it recommends a certain {\em order} of hypotheses to be tested, which is indeed the main point of the present study. Other classic concepts that are recovered naturally from the probabilistic perspective are feature-based indexing and geometric hashing \cite{Lamdan&Wolfson88,Wolfson90}, and feature grouping and perceptual organization \cite{Lowe85,Lowe90,Clemens&Jacobs91}.

To keep the notational load in this article to a minimum and to support ease of reading, we will denote all probability densities by the letter $p$ and indicate each type of density by its arguments. Moreover, we will not introduce symbols for random variables but only for the values they take. It is understood that probability densities become probabilities whenever these values are discrete.

\section{Generative Object Models for Range Data}

Consider the task of recognizing and localizing a number of objects in a scene. More precisely, we want to estimate object parameters $(c,{\bf p})$ from data, where $c \in \bbbn$ is the object's discrete class label and ${\bf p} \in \bbbr^6$ are its pose parameters (rotation and translation). Suppose we use a sensor like a set of stereo cameras or a laser scanner to obtain range-data points ${\bf d} \in \bbbr^3$ of one view of the scene, that is, $2+1/2$ dimensional ($2+1/2$D) range data. A reasonable generative model of the range-data points within the volume ${\cal V}(c,{\bf p})$ of the object $c$ with pose $\bf p$ is given by the conditional probability density
\begin{equation}
\label{p(d|c,p)}
p({\bf d}|c,{\bf p}) = \left\{
\begin{array}{ll}
N(c) \, f\!\left[\phi({\bf d};c,{\bf p})\right] & \mbox{for ${\bf d} \in {\cal S}(c,{\bf p})$,}\\
N(c) \, b & \mbox{for ${\bf d} \in {\cal V}(c,{\bf p}) \backslash {\cal S}(c,{\bf p})$.}
\end{array}
\right.
\end{equation}
The condition is on the object parameters $(c,{\bf p})$. The set ${\cal S}(c,{\bf p}) \subset {\cal V}(c,{\bf p})$ is the region close to visible surfaces of the object $c$ with pose $\bf p$, $b > 0$ is a constant that describes the background of spurious data points, and $N(c)$ is a normalization constant. The surface density of points is described by a function $f$ of the angle $\phi({\bf d};c,{\bf p})$ between the direction of gaze of the sensor and the object's inward surface normal at the (close) surface point $\bf d \in {\cal S}(c,{\bf p})$. The function $f$ takes its maximal value, that we may set to 1, at $\phi = 0$, i.e., on surfaces orthogonal to the sensor's gazing direction.

Four comments are in order. First, the normalization constant $N(c)$ generally depends upon both $c$ {\em and} $\bf p$. However, we here neglect its dependence on the object pose $\bf p$. The consequence is that objects will be harder to detect, if they expose less surface area to the sensor. Second, the extension of the set ${\cal S}(c,{\bf p})$ has to be adapted to the level of noise in the range data. Third, points ${\bf d} \in {\cal S}(c,{\bf p})$ that are close to but not on the surface have assigned surface normals of close surface points. Fourth, given the object parameters $(c,{\bf p})$, the data points ${\bf d} \in {\cal V}(c,{\bf p})$ can be assumed statistically independent with reasonable accuracy.

Our task is to estimate the parameters $c$ and $\bf p$ by optimizing the match between the generative model (\ref{p(d|c,p)}) to the range data $D$. Because of the conditional independence of the data points in ${\cal V}(c,{\bf p})$, this means to maximize the logarithmic likelihood
\begin{equation}
\label{L}
L(c,{\bf p};D) := \sum_{{\bf d} \in D \cap {\cal V}(c,{\bf p})} L(c,{\bf p};{\bf d}) := \sum_{{\bf d} \in D \cap {\cal V}(c,{\bf p})} \ln p({\bf d}|c,{\bf p})
\end{equation}
with respect to $(c,{\bf p}) \in \Omega \subset \bbbn \times \bbbr^6$. A computationally efficient version is obtained by assuming that $\phi \ll 1$, which is true for patches of surface that are approximately orthogonal to the direction of gaze. Fortunately, such parts of the surface contribute most data points; cf.\ (\ref{p(d|c,p)}). Observing that
\begin{equation}
\label{f_expansion}
\ln f(\phi) = -a \, \phi^2 + O\!\left(\phi^4\right) = 2 a \left( \cos \phi - 1 \right) + O\!\left(\phi^4\right) \ ,
\end{equation}
with some constant $a > 0$, we may neglect terms of order $O(\phi^4)$ to obtain
\begin{equation}
L(c,{\bf p};{\bf d}) \approx \left\{
\begin{array}{ll}
2 a \left[{\bf n}({\bf d};c,{\bf p}) \cdot {\bf g} - 1\right] + \ln N(c) & \mbox{for ${\bf d} \in {\cal S}(c,{\bf p})$,}\\
\ln b + \ln N(c) & \mbox{for ${\bf d} \in {\cal V}(c,{\bf p}) \backslash {\cal S}(c,{\bf p})$,}
\end{array}
\right.
\end{equation}
where ${\bf n}({\bf d};c,{\bf p})$ is the object's inward surface-normal vector at the (close) surface point ${\bf d} \in {\cal S}(c,{\bf p})$ and $\bf g$ is the sensor's direction of gaze; both are unit vectors. The constant $\ln b < 0$ is effectively a penalty term for data points that come to lie in the object's interior under the hypothesis $(c,{\bf p})$. Again, discarding the terms $O(\phi^4)$ in (\ref{f_expansion}) is acceptable as data points producing a large error will be rare; cf.\ (\ref{p(d|c,p)}).

\section{Probabilistic Search for Model Matches}

In this section, we derive the probabilistic search algorithm. We first introduce another statistical criterion function of object parameters $(c,{\bf p})$. The new criterion, to be called {\em truncated probability} (TP), defines a sequence of $(c,{\bf p})$-hypotheses. The hypotheses are evaluated by the likelihood (\ref{L}), starting with the most probable $(c,{\bf p})$-value and proceeding to gradually less probable values, as estimated by the TP. The search terminates as soon as $L(c,{\bf p};D)$ is large enough.

We will obtain the TP from a series of steps that simplify from the ideal criterion, the posterior probability of object parameters $(c,{\bf p})$. Although this derivation cannot be regarded as a thoroughly controlled approximation, it will make explicit the underlying assumptions and give some feeling for how far we have to depart from the ideal to arrive at a feasible criterion.

The TP makes use of the probability of the presence of feature values consistent with object parameters $(c,{\bf p})$. It is thus an essential property of the approach to treat features as random variables.

\subsection{The Search Sequence}
\label{TheSearchSequence}

Let us define point features as surface points centered on certain local, isolated surface shapes. Examples of such point features are corners, saddle points, points of locally-extreme surface curvature etc. They are characterized by a pair $(s,{\bf f})$, where $s \in \bbbn$ is the feature's shape-class label and ${\bf f} \in \bbbr^3$ is its location. Consider now a random variable that takes values $(s,{\bf f})$ of point features that are related to the objects sought in the data. Its values are statistically dependent upon the range data $D$. Let us restrict the possible feature values $(s,{\bf f})$ to $s \in \{1,2,\ldots,m\}$ and ${\bf f} \in D$, such that only data points can be feature locations. This is equivalent to setting the feature-value probability to 0 for values $(s,{\bf f})$ with ${\bf f} \notin D$. The restriction is not really correct as we will loose true feature locations between data points. However, it will greatly facilitate the search by limiting features to discrete values that lie close to the true object surfaces and, hence, include highly probable candidates.

Let us introduce the concept of {\em groups} of point features. A feature group is a set $G_g = \{(s_1,{\bf f}_1),(s_2,{\bf f}_2),\ldots,(s_g,{\bf f}_g)\}$ of feature values with ${\bf f}_i \not= {\bf f}_j$ for $i \not= j$. A group $G_g$ hence contains {\em simultaneously possible} values of $g$ features.

The best knowledge we could have about true values of the object parameters $(c,{\bf p})$ is encapsulated in their posterior-probability density given the data $D$, $p(c,{\bf p}|D)$. However, we usually do not know this density, and if we knew it, its maximization would pose a problem similar to our initial one of maximizing the generative model (\ref{L}). We can nonetheless expand it using feature groups,
\begin{equation}
\label{p(c,t|D)_expansion}
p(c,{\bf p}|D) = \sum_{G_g} p(c,{\bf p}|D,G_g) \, p(G_g|D) \ ,\quad
\sum_{G_g} p(G_g|D) = 1 \ .
\end{equation}
The summations are over all possible groups $G_g$ of fixed size $g$. Their enumeration is straightforward but tedious to explicate, so we omit this here. Note that the expansion (\ref{p(c,t|D)_expansion}) is only valid, if we can be sure to find $g$ true feature locations among the data points.

Let us now simplify the density (\ref{p(c,t|D)_expansion}) to the density
\begin{equation}
\label{p_g}
p_g(c,{\bf p};D) := \sum_{G_g} p(c,{\bf p}|G_g) \, p(G_g|D) \ .
\end{equation}
Unlike the posterior density (\ref{p(c,t|D)_expansion}), $p_g$ depends upon the feature-group size $g$. Formally, $p_g$ is a Bayesian belief network with a $g$-feature probability distribution at the intermediate node. Maximization of $p_g$ with respect to $(c,{\bf p})$ is still too difficult a task, as all possible feature-group values contribute to all values of $(c,{\bf p})$-density.

A radically simplifying step thus takes us naturally to the density
\begin{equation}
\label{q_g}
q_g(c,{\bf p};D) := \max_{G_g} p(c,{\bf p}|G_g) \, p(G_g|D) \ ,
\end{equation}
where for each $(c,{\bf p})$ only the largest term in the sum of (\ref{p_g}) contributes. Note that $q_g(c,{\bf p};D) \le p_g(c,{\bf p};D)$ for all $(c,{\bf p})$, such that $q_g$ is not normalized on the set $\Omega$ of object parameters $(c,{\bf p})$. This density will nonetheless be useful for our purpose of guiding a search through $\Omega$, as there only {\em relative} densities matter.

As pointed out above, for all this to make sense, we need $g$ true feature locations among the data points. To be safe, one could be tempted to set $g = 1$. However, the simplifying step from density $p_g$ to density $q_g$ suggests that the latter will be more informative as to the true value of $(c,{\bf p})$, if the sum in (\ref{p_g}) is dominated for high-density values of $(c,{\bf p})$ by only {\em few} terms. Now, less than three point features do not generally define a {\em finite} set of consistent object parameters $(c,{\bf p})$. The density $p(c,{\bf p}|G_g)$ will hence not exhibit a pronounced maximum for $g < 3$. High-density points of $p_g$ arise then from accumulation over {\em many} $g$-feature values, i.e., terms in (\ref{p_g}), and are necessarily lost in $q_g$. Groups of size $g \ge 3$, on the other hand, do define finite sets of consistent $(c,{\bf p})$-values, and $p(c,{\bf p}|G_g)$ has infinite density there. Altogether, feature groups of size $g = 3$ seem to be a good choice for our search procedure; see, however, the discussion in Sect.\ \ref{Conclusion}.

Let us now introduce the logarithmic {\em truncated probability} (TP),
\begin{equation}
\label{TP}
{\rm TP}(c,{\bf p};D) := \lim_{\epsilon \rightarrow 0} \, \ln \int_{S_{\epsilon}({\bf p})} d{\bf p}' \, q_3(c,{\bf p}';D) \ ,
\end{equation}
where $S_{\epsilon}({\bf p})$ is a sphere in pose space centered on $\bf p$ with radius $\epsilon > 0$. The integral and limit are needed to pick up infinities in the density $q_3(c,{\bf p};D)$; see below. The TP is truncated in a double sense: finite contributions from $q_3$ are truncated and $q_3$ itself is obtained from truncating the sum in (\ref{p_g}).

According to the discussion above, it is expected that $(c,{\bf p})$-values of high likelihood (\ref{L}) will mostly yield a high TP (\ref{TP}). Our search thus proceeds by evaluating, in that order,
\begin{multline}
\label{search_seq}
L(c_1,{\bf p}_1;D), L(c_2,{\bf p}_2;D), \ldots \qquad\mbox{with}\\
{\rm TP}(c_1,{\bf p}_1;D) \ge {\rm TP}(c_2,{\bf p}_2;D) \ge \ldots \ , \quad (c_1,{\bf p}_1) = {\rm arg}\max_{(c,{\bf p}) \in \Omega} {\rm TP}(c,{\bf p};D) \ .
\end{multline}
The search stops, as soon as $L(c_k,{\bf p}_k;D) > \Theta$ or all $(c,{\bf p})$-candidates have been evaluated. In the former case, the object identity $c_k$ and object pose ${\bf p}_k$ are returned as the estimates. In the latter case, it is inferred that none of the objects sought is present in the scene. Alternatively, if we know a priori that one of the objects {\em must} be present, we may pick the object parameters that have scored highest under $L$ in the whole sequence. The algorithm will be formulated in more detail in Sect.\ \ref{TheSearchAlg}.

In the density (\ref{q_g}) that guides the search, one of the factors is the density of the object parameters conditioned on the values $G_3$ of a triple of point features. This is explicitly
\begin{equation}
\label{p(c,p|G_3)}
p(c,{\bf p}|G_3) = \left\{
\begin{array}{ll}
\sum_{i=1}^{h(G_3)} \gamma_i(G_3) \, \delta_{c,C_i(G_3)} \, \delta\!\left[{\bf p} - P_i(G_3)\right] &\\
\qquad + \, \left[1 - \sum_{i=1}^{h(G_3)} \gamma_i(G_3)\right] \, \rho_1(c,{\bf p};G_3) & \mbox{for $G_3 \in {\cal C}$,}\\
\rho_2(c,{\bf p};G_3) & \mbox{for $G_3 \notin {\cal C}$.}
\end{array}
\right.
\end{equation}
Here $\delta_{c,c'}$ is the Kronecker delta (elements of the unit matrix) and $\delta({\bf p} - {\bf p}')$ is the Dirac-delta distribution on pose space; $\cal C$ is the set of feature-triple values {\em consistent} with any object hypotheses; $h(G_3)$ is the number of object hypotheses consistent with the feature triple $G_3$; $(C_i(G_3),P_i(G_3))$ are the consistent $(c,{\bf p})$-hypotheses; $\gamma_i(G_3) \in (0,1)$ are probability-weighting factors for the hypotheses. Generally we have
\begin{equation}
\sum_{i=1}^{h(G_3)} \gamma_i(G_3) < 1 \ ,
\end{equation}
which leaves a non-vanishing probability $1 - \sum_i \gamma_i > 0$ that three consistent features do not all belong to one of the objects sought. Accordingly, the density $p(c,{\bf p}|G_3)$ on $\Omega$ contains some finite, spread contribution $\propto \rho_1(c,{\bf p};G_3)$, a ``background'' density, in the consistent-features case. In the inconsistent-features case, the density is similarly spread on $\Omega$ and given by some finite term $\rho_2(c,{\bf p};G_3)$. Clearly, only the values $(C_i,P_i)$ of object parameters where the density (\ref{p(c,p|G_3)}) is infinite will be visited during the search (\ref{search_seq}); cf.\ (\ref{TP}).

The functions $\gamma_i(G_3)$, $C_i(G_3)$, $P_i(G_3)$ are computed by generalized {\em geometric hashing}. Let $G_3 = \{(s_1,{\bf f}_1),(s_2,{\bf f}_2),(s_3,{\bf f}_3)\}$. As key into the hash table we use the pose invariants
\begin{equation}
\label{hash_key}
(s_1,s_2,s_3,||{\bf f}_1-{\bf f}_2||,||{\bf f}_2-{\bf f}_3||,||{\bf f}_3-{\bf f}_1||) \ ,
\end{equation}
appropriately shifted, scaled, and quantized. Here $||\cdot||$ denotes the Euclidean vector norm. The poses $P_i(G_3)$, $i = 1,2,\ldots,h(G_3)$, are computed from matching the triple $G_3$ to $h(G_3)$ model triples from the hash table. Small model distortions that result from an error-tolerant match are removed by orthogonalization of the match-related pose transforms. The weight $\gamma_i(G_3)$ and the object class $C_i(G_3)$ are drawn along with each matched model triple.

The weights $\gamma_i(G_3)$ can be adapted during a training period and during operation of the system in an unsupervised manner. One simply needs to count the number of instances where a feature triple $G_3$ draws a correct set of object parameters $(C_i,P_i)$, i.e., one with $L(C_i,P_i;D) > \Theta$. The adapted $\gamma_i(G_3)$ represent empirical {\em grouping laws} for the point features. Since it will often be the case that $\gamma_i(G_3) \propto 1/h(G_3)$ for all $i = 1,2,\ldots,h(G_3)$, it turns out that hypotheses drawn by feature triples {\em less common} among the sought objects tend to be visited {\em before} the ones drawn by more common triples during the search (\ref{search_seq}).

In order to fully specify the TP (\ref{TP}) and formulate our search algorithm, it remains to establish a model for the feature-value probabilities $p(G_3|D)$ given the data $D$; cf.\ (\ref{q_g}). This is the subject of the next section.

\subsection{Inferring Local Surface Shape from Point-Relation Densities}

In principle, a generative model of range data on local surface shapes $s$ can be built analogous to (\ref{p(d|c,p)}). Thus, we may construct a conditional probability density $p(D|s,{\bf f},{\bf r})$, where ${\bf r} \in \bbbr^3$ are the rotational parameters of the local surface patch represented by the point feature $(s,{\bf f})$. During recognition, however, it would then be necessary to optimize the parameters $\bf r$ for each $s \in \{1,2,\ldots,m\}$ and ${\bf f} \in D$, in order to obtain the likelihood of each feature value $(s,{\bf f})$. Besides the huge overhead of doing so without being interested in an estimate for $\bf r$, this procedure would confront us with a problem similar to our original one of optimizing $p(D|c,{\bf p})$. On the other hand, the density $p(D|s,{\bf f})$ is an infeasible model because of high-order correlations between the data points $D$. Neglecting these correlations would throw out all information on surface shape $s$.

The strategy we pursue here is to capture some of the informative correlations between data points $D$ in a family of new representations $T({\bf c}) = \{\Delta_1,\Delta_2,\ldots,\Delta_N\}$, ${\bf c} \in \bbbr^3$. Each $T({\bf c})$ represents the geometry of the data $D$ relative to the point ${\bf c}$, and each $\Delta_i$ represents geometric relations between multiple data points. A reasonable statistical model is then obtained by neglecting correlations between the $\Delta_i$.

\subsubsection{Learning Point-Relation Densities.}
\label{LearningPRDs}

\begin{figure}
\begin{minipage}{0.20\textwidth}
\psfrag{d1}{${\bf d}_1$}
\psfrag{d2}{${\bf d}_2$}
\psfrag{d3}{${\bf d}_3$}
\psfrag{c}{$\bf c$}
\psfrag{r}{$r$}
\psfrag{d}{$d$}
\psfrag{a}{$a$}
\leftline{\includegraphics[width=\textwidth]{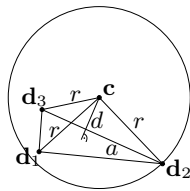}}
\end{minipage}
\begin{minipage}{0.79\textwidth}
\caption{Transformation of range data points to a tetrahedron representation. For an inspection point $\bf c$, the 3-tuples $(r,d,a)$ are collected and normalized [cf.\ (\ref{Delta})] for each triple of data points ${\bf d}_1, {\bf d}_2, {\bf d}_3 \in D$ with (approximately) equal distance $r \in (0,R)$ from $\bf c$.
\label{tetrahedron_rep}}
\end{minipage}
\end{figure}

The point relations we are going to exploit here are between four data points, that is, tetrahedron geometries, where three of the points are selected from a spherical neighborhood of the fourth. For four points ${\bf x}_1, {\bf x}_2, {\bf x}_3, {\bf c} \in \bbbr^3$ we define the map
\begin{equation}
\label{Delta}
\left(\begin{array}{c} {\bf x}_1 \\ {\bf x}_2 \\ {\bf x}_3 \end{array}\right)
\mapsto \Delta\!\left({\bf c}; {\bf x}_1, {\bf x}_2, {\bf x}_3\right)
:= \left(\begin{array}{c}
r/R\\
d \Big/ \sqrt{r^2 - \frac{4 a}{3 \sqrt{3}}}\\
4 a/3 \sqrt{3} (r^2 - d^2)
\end{array}\right)
\in [0,1] \times [-1,1] \times [0,1] \ ,
\end{equation}
where $r$ is the mean distance of the center $\bf c$ to ${\bf x}_1, {\bf x}_2, {\bf x}_3$, $R$ is the radius of the spherical neighborhood of $\bf c$, $d$ is the (signed) distance of $\bf c$ to the plane defined by ${\bf x}_1, {\bf x}_2, {\bf x}_3$, and $a$ is the area of the triangle they define\footnote{The quantity introduced in (\ref{Delta}) will be denoted by $\Delta({\bf c}; {\bf x}_1, {\bf x}_2, {\bf x}_3)$, $\Delta({\bf c})$, or simply $\Delta$ to indicate its dependence on points as needed.}. Explicitly,
\begin{eqnarray}
r & = & \frac{1}{3} \sum_{i=1}^3 ||{\bf x}_i - {\bf c}|| \ ,\\
\label{d}
d & = & {\rm sgn}\!\left[{\bf g} \cdot \left({\bf x}_1 - {\bf c}\right)\right] \, \frac{||\left({\bf x}_1 - {\bf c}\right) \cdot \left[\left({\bf x}_2 - {\bf x}_1\right) \times \left({\bf x}_3 - {\bf x}_1\right)\right]||}{||\left({\bf x}_2 - {\bf x}_1\right) \times \left({\bf x}_3 - {\bf x}_1\right)||} \ ,\\
a & = & \frac{1}{2} \, ||\left({\bf x}_2 - {\bf x}_1\right) \times \left({\bf x}_3 - {\bf x}_1\right)|| \ .
\end{eqnarray}
As seen in (\ref{d}), the sign of $d$ is determined by the direction of gaze $\bf g$. We can now define the family of {\em tetrahedron representations} $T({\bf c})$ by
\begin{multline}
\label{tetrahedron_rep_def}
T({\bf c}) := \Big\{ \Delta({\bf c}; {\bf d}_1, {\bf d}_2, {\bf d}_3) \, \Big| \, \{{\bf d}_1, {\bf d}_2, {\bf d}_3\} \subseteq D
\wedge \max_{i=1,2,3} ||{\bf d}_i - {\bf c}|| < R\\
\wedge \max_{i=1,2,3} ||{\bf d}_i - {\bf c}|| - \min_{i=1,2,3} ||{\bf d}_i - {\bf c}|| < \epsilon \Big\} \ .
\end{multline}
The parameter $\epsilon > 0$ sets a small tolerance for differences in distance to $\bf c$ among each data-point triple ${\bf d}_1, {\bf d}_2, {\bf d}_3$, that is, ideally all three points have the same distance. In practice, the tolerance $\epsilon$ has to be adapted to the density of range-data points $D$ to obtain enough $\Delta$-samples in each $T({\bf c})$. Figure \ref{tetrahedron_rep} depicts the geometry of the tetrahedron representations.

One of the nice properties of the tetrahedron representation (\ref{tetrahedron_rep_def}) is that for any surface of revolution we get at its symmetry point $\bf c$ samples $\Delta \in T({\bf c})$ that are confined to a surface in $\Delta$-space characteristic of the original surface shape. For surfaces of revolution, the distinctiveness and low entropy of the distribution of range data $D$ sampled from a surface in 3D is thus completely preserved in the tetrahedron representation. For more general surfaces, it is therefore expected that a high mutual information between $\Delta$-samples and shape classes can be achieved.

Our goal is to estimate {\em point-relation densities} for each shape class $s = 1,2,\ldots,m$. Samples are taken from tetrahedron representations centered on feature locations $\bf f$ from a training set ${\cal F}_s$, that is, from $\cup_{{\bf f} \in {\cal F}_s} T({\bf f})$. Moreover, we add a non-feature class $s = 0$ that lumps together all the shapes we are not interested in. The result are the shape-conditioned densities $p(\Delta|s)$ for $\Delta \in [0,1] \times [-1,1] \times [0,1]$ and $s = 0,1,2,\ldots,m$. In particular, $p(\Delta|s) = p(\Delta|s,{\bf f})$ for $\Delta \in T({\bf f})$.

Estimation of $p(\Delta|s)$ is made simple by the fact that we get $O(n^3)$, i.e., a lot of samples of $\Delta$ for each feature location $\bf f$ with $n$ data points within a distance $R$. For our application, it is sufficient to count $\Delta$-samples in bins to estimate probabilities $p(\Delta|s)$ for discrete $\Delta$-events as normalized frequencies.

\subsubsection{Inferring Local Surface Shape.}
\label{InferringLSS}

Let the features' prior probabilities be $p(s)$ for $s = 1,2,\ldots,m$, that is, $p(s) \in (0,1)$ is the probability that any given data point from $D$ is a feature location of type $s$. The feature priors are virtually impossible to know for general scenes. We do know, however, that $p(s) \ll 1$ for $s = 1,2,\ldots,m$, i.e., the overwhelming majority of data points are not feature locations. This must be so for a useful dictionary of point features. It thus makes sense to expand the logarithm of the shapes' posterior probabilities, given $l$ samples $\Delta_1({\bf f}),\Delta_2({\bf f}),\ldots,\Delta_l({\bf f}) \in T({\bf f})$, ${\bf f} \in D$,
\begin{multline}
\label{lnp(s|T)_expansion}
\ln p\!\left[s|\Delta_1({\bf f}),\Delta_2({\bf f}),\ldots,\Delta_l({\bf f})\right] = \ln p(s)
+ \sum_{i=1}^l \left\{\ln p\!\left[\Delta_i({\bf f})|s\right] - \ln p\!\left[\Delta_i({\bf f})|0\right]\right\}\\
+ \, O\!\left[p(1),p(2),\ldots,p(m)\right] \ ,
\end{multline}
and neglect the terms $O[p(1),p(2),\ldots,p(m)]$. Remember that $p(\Delta|0)$ is the $\Delta$-density of the non-feature class. The expression (\ref{lnp(s|T)_expansion}) neglects correlations between the $\Delta_i({\bf f}) \in T({\bf f})$.

The sample length $l$ for each representation $T({\bf f})$ will usually be $l \ll |T({\bf f})|$, the cardinality of the set $T({\bf f})$. It is in fact crucial that we do not have to generate the complete representations $T({\bf f})$ for recognition, as this would require $O(n^3)$ time for $n$ data points within a distance $R$ from $\bf f$. Instead, it is possible to draw a random, unbiased sub-sample of $T({\bf f})$ in $O(n)$ and $O(l)$ time.

The first term in (\ref{lnp(s|T)_expansion}) can be split into
\begin{equation}
\label{lnp(s)}
\ln p(s) = \ln q(s) + \ln\sum_{s'=1}^m p(s') \ ,
\end{equation}
where $q(s) \in (0,1)$ are the relative frequencies of shape classes $s = 1,2,\ldots,m$. Now, these frequencies do not differ between features over many orders of magnitude for a useful dictionary of point features\footnote{A feature that is more than a hundred times less frequent than the others should usually be dropped for computational efficiency.}. The dependence of (\ref{lnp(s)}) on the shape class $s$ is thus negligible compared to the sum in (\ref{lnp(s|T)_expansion}) for reasonable sample lengths $l \gg 1$ (several tens to hundreds). The first term in (\ref{lnp(s|T)_expansion}) may hence be disregarded, and we are left with the feature's log-probability
\begin{multline}
\label{Phi}
\Phi(s,{\bf f};D) := \ln p\!\left[s|\Delta_1({\bf f}),\Delta_2({\bf f}),\ldots,\Delta_l({\bf f})\right] + {\rm const.}\\
\approx \sum_{i=1}^l \left\{\ln p\!\left[\Delta_i({\bf f})|s\right] - \ln p\!\left[\Delta_i({\bf f})|0\right]\right\} \ ,
\end{multline}
that is, a log-likelihood ratio.

\subsection{The Search Algorithm}
\label{TheSearchAlg}

We still need to determine the TP (\ref{TP}) for guiding the search (\ref{search_seq}). Let again $G_3 = \{(s_1,{\bf f}_1),(s_2,{\bf f}_2),(s_3,{\bf f}_3)\}$. It is reasonable to assume statistical independence of the point features in $G_3$, as long as there are many different objects in the world that have these features\footnote{These objects may or may not be in our modeled set.}. We hence model the feature-triples' posterior probability $p(G_3|D)$ in (\ref{q_g}) as
\begin{equation}
p(G_3|D) = \prod_{i=1}^3 p\!\left[s_i|\Delta_1({\bf f}_i),\Delta_2({\bf f}_i),\ldots,\Delta_l({\bf f}_i)\right] \propto \exp\left[ \sum_{i=1}^3 \Phi\!\left(s_i,{\bf f}_i;D\right) \right] \ .
\end{equation}
The TP can then be expressed as
\begin{equation}
\label{TP_explicit}
{\rm TP}(c,{\bf p};D) = \max_{G_3 \in {\cal C}} \left[ \ln \sum_{i=1}^{h(G_3)} \gamma_i(G_3) \, \delta_{c,C_i(G_3)} \, \delta_{{\bf p},P_i(G_3)} + \sum_{i=1}^3 \Phi\!\left(s_i,{\bf f}_i;D\right) \right] \ ,
\end{equation}
up to additive constants. The first term under the max-operation is the contribution from feature grouping\footnote{The factor $\delta_{{\bf p},P_i(G_3)}$ is an idealization. Since the pose parameters $\bf p$ are continuous, the hashing with the key (\ref{hash_key}) is necessarily error tolerant. The factor is then to be replaced by an integral $\int_{{\cal P}_i(G_3)} d{\bf p}' \, \delta({\bf p} - {\bf p}')$ for a set ${\cal P}_i(G_3)$ of pose parameters.}, the second from single features. In particular, we get
\begin{multline}
{\rm TP}(c,{\bf p};D) = -\infty \mbox{ for}\\
(c,{\bf p}) \notin \left\{\left(C_i(G_3),P_i(G_3)\right) \, \big| \, G_3 \in {\cal C} , \, i \in \{1,2,\ldots,h(G_3)\} \right\} \ ,
\end{multline}
that is, for $(c,{\bf p})$-values not suggested by any feature triple $G_3$.

We are now prepared to formulate the search algorithm. The following is not meant to specify a flow of processing, but rather a logic sequence of operations.
\begin{enumerate}
\item\label{fcollecting} From all possible feature values $(s,{\bf f}) \in \{1,2,\ldots,m\} \times D$, select as feature candidates the values with log-probability $\Phi(s,{\bf f};D) > \Xi$.
\item\label{hashing} From the feature candidates, generate all triples $G_3$. Use their keys (\ref{hash_key}) into a hash table to find the associated object hypotheses $(C_i(G_3),P_i(G_3))$ and their weights $\gamma_i(G_3)$ for $i = 1,2,\ldots,h(G_3)$.
\item\label{scoring} For the object hypotheses obtained, compute the TPs.
\item\label{evaluation} Evaluate the object hypotheses by the likelihood $L$ in the order of decreasing TP, starting with the hypothesis scoring highest under TP. Skip duplicate hypotheses. Stop when $L(c,{\bf p};D) > \Theta$ for a hypothesis $(c,{\bf p})$.
\item\label{result} Return the last evaluated object parameters $(c,{\bf p})$ as the result, if $L(c,{\bf p};D) > \Theta$. Else conclude that there is none of the sought objects in the data.
\end{enumerate}
Some comments are in order.
\begin{itemize}
\item Comment on step \ref{fcollecting}: The restriction to the most probable feature values by introduction of the threshold $\Xi$ is optional. It is a powerful means, however, to radically speed up the algorithm, if there are many range data points $D$. If the probability measure on the features is reliable and $\Xi$ is adjusted conservatively, the risk of missing a good set of object parameters is very small; see Sect.\ \ref{LSRecognition} below.
\item Comment on step \ref{hashing}: Feature triples $G_3 \notin {\cal C}$, that is, inconsistent triples, are discarded by the hashing process.
\item Comment on step \ref{scoring}: Computation of TP is cheap as the hashed weights $\gamma_i$ from step \ref{hashing} and the feature's log-probabilities $\Phi$ from step \ref{fcollecting} can be used; cf.\ (\ref{TP_explicit}).
\item Comment on step \ref{evaluation}: Duplicate hypotheses may be very unlikely, but this depends on when two hypotheses are considered equal. In any case, skipping them realizes the $\max_{G_3 \in {\cal C}}$-operation in the TP (\ref{TP_explicit}), ensuring that each hypothesis is drawn only by the feature triple which makes it most probable.
\item Comment on step \ref{result}: If the likelihood $L$ does not reach the threshold $\Theta$ but we know a priori that at least one of the objects must be present in the scene, the algorithm may return the object parameters $(c,{\bf p})$ that have scored highest under $L$.
\end{itemize}

A complete scene interpretation often involves more than one recognized set of object parameters $(c,{\bf p})$. A simultaneous search for several objects is possible within the current framework, although it will be computationally very costly. It requires evaluation of TPs and likelihoods
\begin{equation}
{\rm TP}(c_1,{\bf p}_1,c_2,{\bf p}_2,\ldots;D) \ ,\quad L(c_1,{\bf p}_1,c_2,{\bf p}_2,\ldots;D)
\end{equation}
for a combinatorially enlarged set of object parameters $(c_1,{\bf p}_1,c_2,{\bf p}_2,\ldots)$. A much cheaper, although less reliable, alternative is sequential search, where the data belonging to a recognized object $(c,{\bf p})$ are removed prior to the algorithm's restart.

\section{Experiments}
\label{Experiments}

We have performed two types of experiment. One tests the quality of the measure (\ref{Phi}) of the features' posterior probability. The other probes the capabilities of the complete algorithm for object segmentation and recognition. All experiments were run on stereo data obtained from a three-camera device (the `Triclops Stereo Vision System', Point Grey Research Inc.). The stereo data were calculated from three $120 \times 160$-pixel images by a standard least-sum-of-pixel-differences algorithm for correspondence search. The data were accordingly of a rather low resolution. They were moreover noisy, contained a lot of artifacts and outliers, and even visible surfaces lacked large regions of data points, as is not uncommon for stereo data; see Figs.\ \ref{obj_recog_scene1} and \ref{obj_recog_scene2}.

\subsection{Local Shape Recognition}
\label{LSRecognition}

Only correct feature values are able to draw correct object hypotheses $(c,{\bf p})$, except from accidental, unprobable events. For an acceptable search length it is, therefore, crucial that correct feature values yield a high log-probability $\Phi$; cf.\ (\ref{Phi}) and (\ref{TP_explicit}). This is even more crucial, if feature candidates are preselected by imposing a threshold $\Phi(s,{\bf f};D) > \Xi$; cf.\ step \ref{fcollecting} of the algorithm in Sect.\ \ref{TheSearchAlg}.

As point features, we chose convex and concave rectangular corners. The training set contained 221 feature locations that produced 466,485 $\Delta$-samples for the densities $p(\Delta|s)$, $s \in \{\mbox{``convex corner''},\mbox{``concave corner''}\}$. For the non-feature class $s = 0$ we obtained 9,623,073 $\Delta$-samples from several thousand non-feature locations (mostly on planes, also some edges). The parameters for $\Delta$ sampling were $R = 15$ mm and $\epsilon = 1$ mm; cf.\ Sect.\ \ref{LearningPRDs}. The $\Delta$-samples were counted in $15 \times 20 \times 10$ bins, which corresponds to the discretization used for the $\Delta$-space $[0,1] \times [-1,1] \times [0,1]$.

Feature-recognition performance was evaluated on 10 scenes that contained (i) an object with planar surface segments, edges, saddle points, and altogether 65 convex and concave corner points (feature locations); (ii) a piece of loosely folded cloth that contributed a lot of potential distractor shapes. The sample length was $l = 50$ $\Delta$-samples; cf.\ Sect.\ \ref{InferringLSS}. Since drawing $\Delta$-samples is a stochastic process, we let the program run 10 times on each scene with different seeds for the random-number generator to obtain a more significant statistics.

\begin{figure}
\psfrag{Phi-score}{$\Phi$-score}
\psfrag{C-score}{$C$-score}
\centerline{\includegraphics[width=0.7\textwidth]{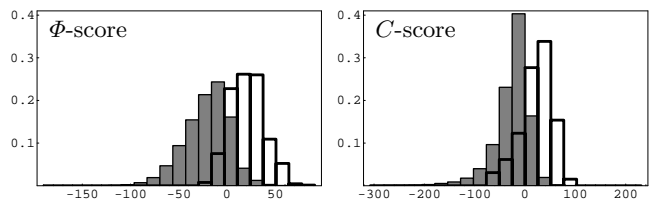}}
\caption{Score distribution of true (transparent bars, thick lines) and all possible (filled bars, thin lines) feature values as produced by the $\Phi$-score (left) and the $C$-score (right). Horizontally in each plot extends the score, vertically the normalized frequency of feature values.
\label{fscores}}
\end{figure}

\begin{figure}
\psfrag{Phi-rank}{$\Phi$-rank}
\psfrag{C-rank}{$C$-rank}
\centerline{\includegraphics[width=0.7\textwidth]{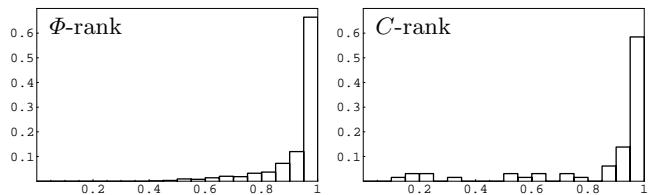}}
\caption{Rank distribution of true feature values among all possible feature values as produced by the $\Phi$-score (left) and the $C$-score (right); rank 1 is the highest score in each data set, rank 0 the lowest score. Horizontally in each plot extends the rank, vertically the normalized frequency of true feature values.
\label{franks}}
\end{figure}

Corner points are particularly interesting as test features, since they allow for comparison of the proposed log-probability measure (\ref{Phi}) with the classic method of curvature estimation. Let $c_1({\bf f};D)$ and $c_2({\bf f};D)$ be the principal curvatures of a second-order polynomial that is least-squares fitted to the data points from $D$ in the $R$-sphere around the point $\bf f$. Because the corners are the maximally curved parabolic shapes in the scenes, it makes sense to use the curvature measures
\begin{eqnarray}
C(s = \mbox{``convex corner''},{\bf f};D) & := & \min\!\left[c_1({\bf f};D),c_2({\bf f};D)\right] \ ,\\
C(s = \mbox{``concave corner''},{\bf f};D) & := & \min\!\left[-c_1({\bf f};D),-c_2({\bf f};D)\right] \ ,
\end{eqnarray}
as alternative measures of ``cornerness'' of the point $\bf f$ for convex and concave shapes, respectively. We thus compare $\Phi(s,{\bf f};D)$ as defined in (\ref{Phi}) to $C(s,{\bf f};D)$ on true and false feature values
\begin{equation}
(s,{\bf f}) \in \{\mbox{``convex corner''},\mbox{``concave corner''}\} \times D \ .
\end{equation}
In Fig.\ \ref{fscores} we show the distribution of $\Phi$- and $C$-scores for the 65 true feature values in relation to their distribution for all possible feature values, which are more than 100,000. As can be seen, the $\Phi$- and $C$-distributions for the population of true values are both distinct from, but overlapping with the $\Phi$- and $C$-distributions for the entire population of feature values. A qualitative difference between $\Phi$- and $C$-scoring of feature values is more obvious in the {\em ranking} they produce. In Fig.\ \ref{franks} we show the distribution of $\Phi$- and $C$-ranks of the true feature values among all feature values, with 1 being the first rank, i.e., highest $\Phi$/$C$-score, and 0 the last. Both $\Phi$- and $C$-ranks of true values are most frequently found close to 1 and gradually less frequently at lower ranks. There are almost no true features $\Phi$-ranked below 0.6. The $C$-ranks, in contrast, seem to scatter uniformly in $[0,1]$ for part of the true features. These features score under $C$ as poorly as the non-features. If using the $C$-score, they would in practice not be available for object recognition, as they would only be selected to draw an object hypothesis after hundreds to thousands of false feature values.

The result of the comparison comes not as a complete surprise. Some kind of surface-fitting procedure is necessary for estimation of the principal curvatures \cite{Besl&Jain85a,Besl&Jain85b,Stokely&YouWu92}. If not a lot of care is invested, fitting to data will always be very sensitive to outliers and artifacts, of which there are many in typical stereo data. A robust fitting procedure, however, has to incorporate in some way a statistical model of the data that accounts for its outliers and artifacts. We here propose to go all the way to a statistical model, the point-relation densities, and to do without any fitting for local shape recognition.

\subsection{Object Segmentation and Recognition}

We have tested segmentation and recognition performance of the proposed algorithm on two objects. One is a cube from which 8 smaller cubes are removed at its corners; the other is a subset of this object. These two objects can be assembled to create challenging segmentation and recognition scenarios.

The test data were taken from 50 scenes that contained the two objects to be recognized, and additional distractor objects in some of them. A single data set consisted of roughly 5,000 to 10,000 3D points. As an acceptable accuracy for recognition of an object's pose we considered what is needed for grasping with a hand (robot or human), i.e., less than 3 degrees of angular and less than 5 mm of translational misalignment. For a fixed set of parameters of the search algorithm, in all but 3 cases the result was acceptable in this sense. Processing times ranged roughly between 1 and 50 seconds, staying mostly well below 5 seconds, on a Sun UltraSPARC-III workstation (750 MHz).

\begin{figure}
\centerline{\includegraphics[width=\textwidth]{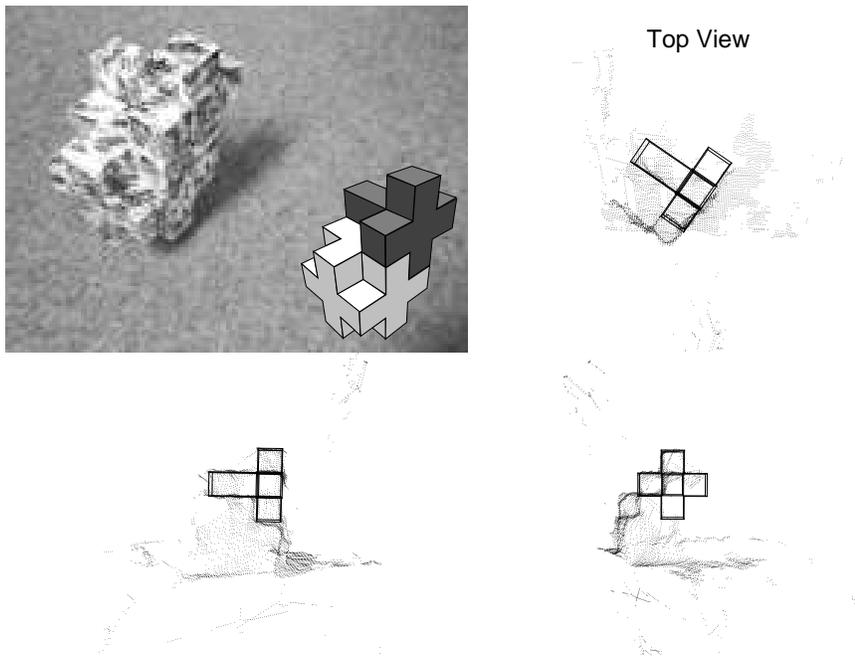}}
\caption{Example scene with the two objects we used for evaluation of segmentation and recognition performance. The smaller object is stacked on the larger as drawn near the center of the figure. Shown are one of the camera images and three orthogonal views of the stereo-data set. The object pose recognized first is outlined in the data. The length of both objects' longest edges is 13 cm. The recognition time was 1.2 seconds.
\label{obj_recog_scene1}}
\end{figure}

\begin{figure}
\centerline{\includegraphics[width=\textwidth]{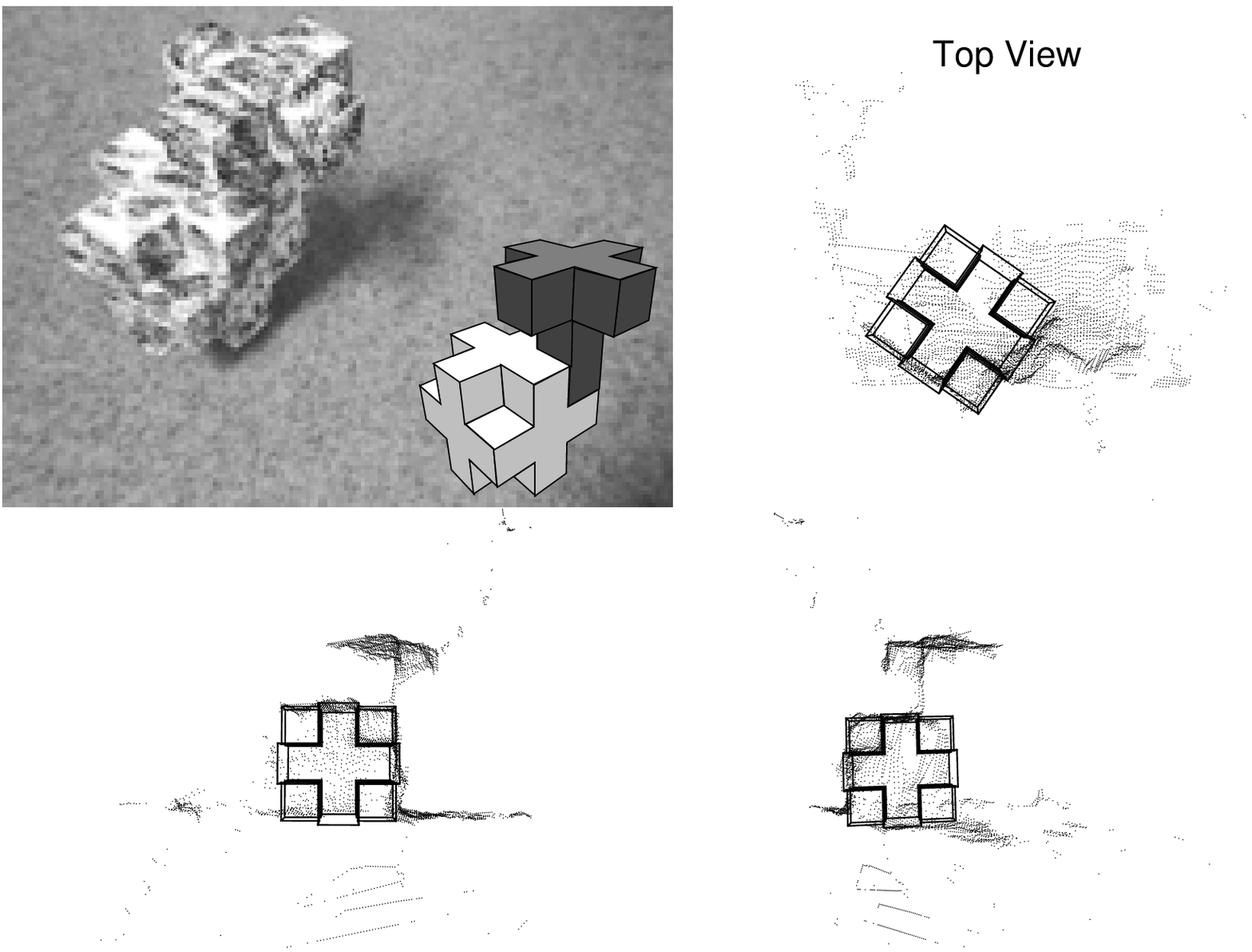}}
\caption{Example scene with the two objects we used for evaluation of segmentation and recognition performance; cf.\ Fig.\ \ref{obj_recog_scene1}. The recognition time was 1.5 seconds.
\label{obj_recog_scene2}}
\end{figure}

In Figs.\ \ref{obj_recog_scene1} and \ref{obj_recog_scene2} we show examples of scenes, camera images, range data, and recognition results. Edges drawn in the data outline the object pose recognized first. The other object can be recognized after deletion of data points belonging to the first object.

\section{Conclusion}
\label{Conclusion}

We have derived from a probabilistic perspective a new variant of hypothesize-and-test algorithm for object recognition. A critical element of any such algorithm is the ordered sequence of hypotheses that is tested. The ordered sequence that is generated by our algorithm is inferred from a statistical criterion, here called the {\em truncated probability} (TP) of object parameters. As a key component of the truncated probability, we have introduced {\em point-relation densities}, from which we obtain an estimate of posterior probability for surface shape given range data.

One of the strengths of the algorithm, as demonstrated in experiments, is its very high degree of robustness to noise and artifacts in the data. Raw stereo-data points of low quality are sufficient for many recognition tasks. Such data are fast and cheap to obtain and are intrinsically invariant to changes in illumination.

We have argued in Sect.\ \ref{TheSearchSequence} that feature groups of size $g = 3$, i.e., minimal groups, are a good choice for guiding the search for model matches. However, if we can be sure that more than three true object features are represented in the data, it may be worth considering larger groups. For $g > 3$, the density (\ref{p(c,p|G_3)}) that enters the TP (\ref{TP}) looks more complicated. In particular, hash tables of higher dimension are needed to accommodate the grouping laws, i.e., the weights $\gamma(G_g)$. If we do without the grouping laws, using larger groups is equivalent to having more than just the largest term of the density (\ref{p_g}) contributing to the TP; cf.\ (\ref{q_g}). In the limit of large feature groups, the approach then resembles generalized-Hough-transform and pose-clustering techniques \cite{Ballard81,Stockman87,Moss_etal99}.

One avenue of research that is suggested here is for generalizations of point-relation densities. Alternative representations of range data have to be explored for learning posterior-probability estimates for various surface shapes.

\end{document}